\documentclass{article}

\usepackage{spconf,amsmath,graphicx,hyperref}
\usepackage{multirow}
\usepackage{array}
\usepackage{amssymb}

\usepackage{enumitem}
\setlist{nosep} 
\usepackage[table]{xcolor}
\usepackage{booktabs}
\usepackage{comment}

\DeclareRobustCommand\onedot{\futurelet\@let@token\@onedot}
\def\onedot{.} 
\def\eg{\emph{e.g}\onedot, }

\title{Explicit Time-Frequency Dynamics for Skeleton-Based Gait Recognition}
\name{Seoyeon Ko$^1$, Yeojin Song$^1$, Egene Chung$^1$, Luca Quagliato$^2$, Taeyong Lee$^1$, Junhyug Noh$^1$
}
\address{
$^1$Ewha Womans University, Seoul, Korea\\
$^2$University of Trento, Trento TN, Italy
}
\begin{document}
%
\maketitle
\begin{abstract}
Skeleton-based gait recognizers excel at modeling spatial configurations but often underuse \emph{explicit} motion dynamics that are crucial under appearance changes. We introduce a plug-and-play \emph{Wavelet Feature Stream} that augments any skeleton backbone with time--frequency dynamics of joint velocities. Concretely, per-joint velocity sequences are transformed by the continuous wavelet transform (CWT) into multi-scale scalograms, from which a lightweight multi-scale CNN learns discriminative dynamic cues. The resulting descriptor is fused with the backbone representation for classification, requiring no changes to the backbone architecture or additional supervision. Across CASIA-B, the proposed stream delivers consistent gains on strong skeleton backbones (\eg GaitMixer, GaitFormer, GaitGraph) and establishes a new skeleton-based state of the art when attached to GaitMixer. The improvements are especially pronounced under covariate shifts such as carrying bags (BG) and wearing coats (CL), highlighting the complementarity of explicit time--frequency modeling and standard spatio--temporal encoders.
\end{abstract}
\section{Introduction}
\label{sec:intro}
Gait recognition identifies individuals from their walking patterns and is attractive for its non-invasive, long-range nature. Methods are commonly grouped into \emph{appearance-based} and \emph{skeleton-based} approaches.
Appearance-based systems learn from silhouettes~\cite{silhouette, 2006GEI} or pixel intensities~\cite{castro2020multimodal} and currently report state-of-the-art results on CASIA-B (\eg 3DLocal~\cite{huang20213d}).
However, their accuracy often degrades under covariates such as clothing changes (CL) or carrying bags (BG).
Skeleton-based approaches, by contrast, operate on joint coordinates and are naturally robust to texture/appearance variations, but they rely on pose-estimation quality and typically model motion dynamics only implicitly.

Recent skeleton models (\eg GaitMixer, GaitFormer, GaitGraph) have closed much of the gap by improving spatio-temporal modeling.
Yet a persistent limitation remains: \emph{explicit} treatment of motion dynamics -- how joints evolve over time in multiple frequency bands -- is underexplored, even though dynamics constitute a complementary and discriminative component of gait.

\begin{figure}[t!]
    \centering
    \includegraphics[width=\columnwidth]{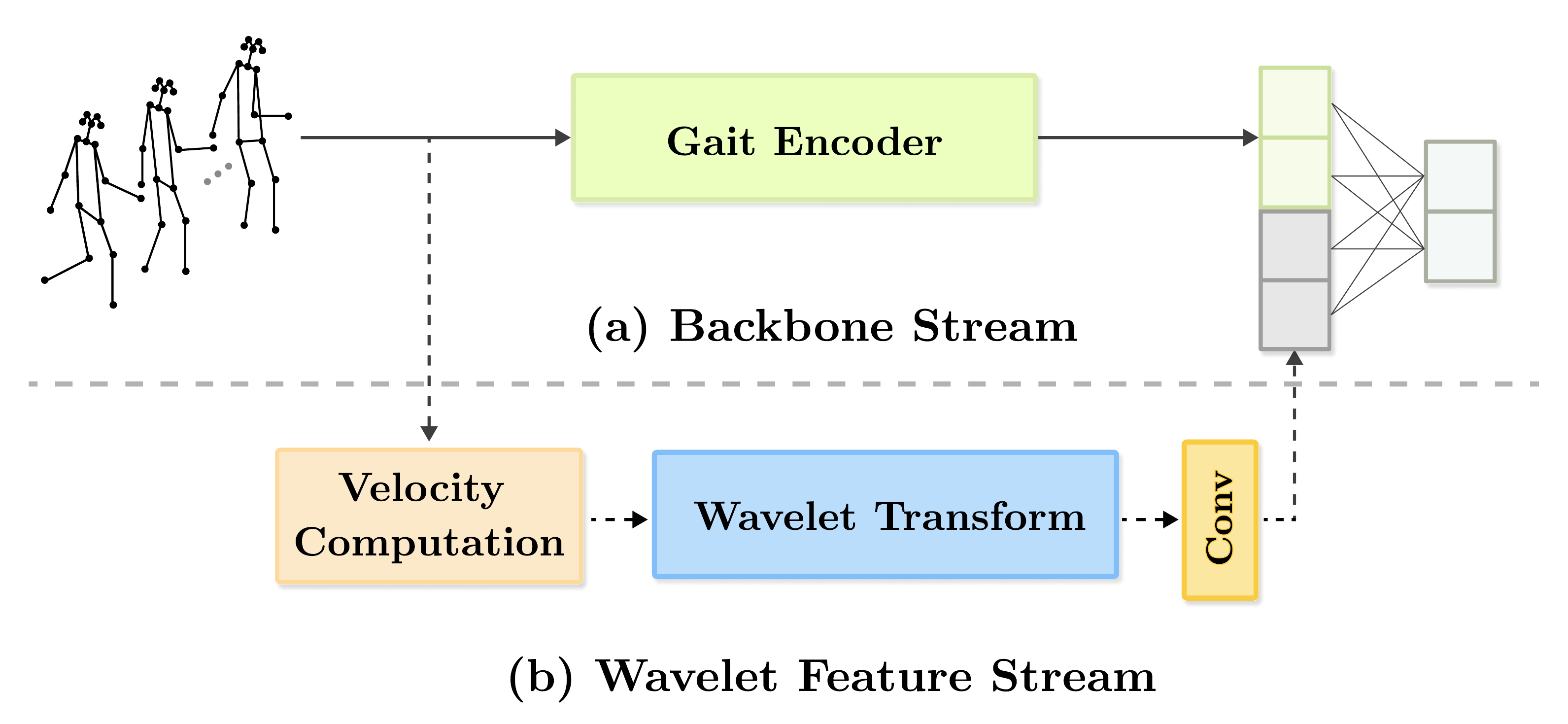}
    \caption{Overview of the proposed plug-and-play Wavelet Feature Stream attached to a skeleton backbone. Per-joint velocities $\rightarrow$ CWT scalograms $\rightarrow$ multi-scale CNN $\rightarrow$ fusion with backbone features.}
    \label{fig:overview}
\end{figure}

We address this gap with a \emph{Wavelet Feature Stream} that encodes joint-velocity signals in the time-frequency domain and augments any skeleton backbone in a plug-and-play manner.
The key idea is to transform per-joint velocities using the continuous wavelet transform (CWT) to obtain multi-scale scalograms, from which a lightweight multi-scale CNN learns discriminative dynamic cues.
The resulting descriptor is fused with the backbone representation for classification (Figure~\ref{fig:overview}).
Wavelet representations are particularly suitable for non-stationary motion signals, offering localized, multi-resolution analysis that complements conventional spatio-temporal encoders.

In summary, our contributions are:
\begin{itemize}[leftmargin=*]
    \item We propose a \emph{plug-and-play Wavelet Feature Stream} that injects \emph{explicit} time-frequency dynamics of joint velocities via CWT into skeleton-based gait recognition.
    \item On CASIA-B, attaching our stream to GaitMixer establishes a new \emph{skeleton-based} state-of-the-art (SOTA) and yields \emph{consistent} improvements across multiple backbones, with particularly notable gains under BG/CL covariates.
    \item Under the most challenging coat condition (CL), our proposed skeleton model is the first to exceed the performance of existing \emph{appearance-based} SOTA models, proving the efficiency and potential of skeleton information.
\end{itemize}

\begin{figure*}[t!]
    \centering 
    \includegraphics[width=0.95\textwidth]{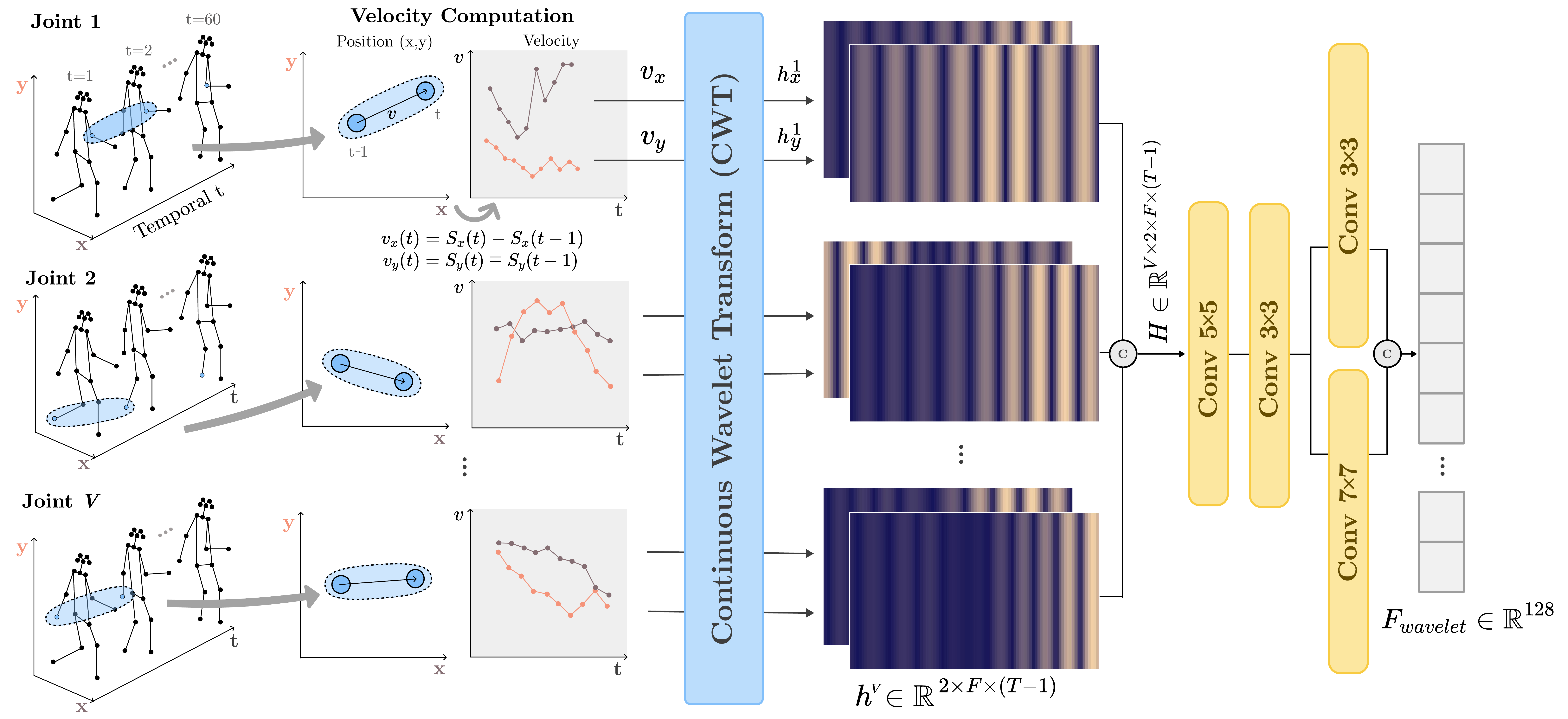}
    \caption{\textbf{Wavelet Feature Stream.} Given per-joint velocities from skeleton sequences, we apply CWT to obtain time--frequency maps (scalograms) and learn multi-scale dynamics via a lightweight CNN. The resulting descriptor is fused with a backbone feature for classification.}
    \label{fig:wavelet} 
\end{figure*}

\section{Related Work}
\label{sec:relatedwork}

Gait recognition identifies individuals from walking patterns and is typically approached via \emph{appearance-based} or \emph{skeleton-based} methods. Appearance models currently lead on CASIA-B, while skeleton models have rapidly narrowed the gap with improved spatio--temporal modeling.

\noindent\textbf{Appearance-based.}
Silhouette/pixel methods such as GaitNet~\cite{song2019gaitnet}, GaitSet~\cite{chao2021gaitset}, GaitPart~\cite{fan2020gaitpart}, and 3DLocal~\cite{huang20213d} learn strong subject representations and report state-of-the-art results. Their limitations include reliance on silhouette quality and the loss of fine visual cues for disambiguating similar body shapes.

\noindent\textbf{Skeleton-based.}
Joint-graph methods including PoseGait~\cite{liao2020model}, GaitGraph/GaitGraph2~\cite{teepe2021gaitgraph,teepe2022towards}, and GaitMixer~\cite{pinyoanuntapong2023gaitmixer} model kinematic structure and are naturally robust to appearance changes. However, they depend on pose-estimation accuracy and typically underuse \emph{explicit} motion dynamics. 
\emph{Our work} complements such backbones by injecting a continuous wavelet transform (CWT) based, explicit time--frequency representation of joint velocities~\cite{sivakumar2021joint, ji2019appropriate} and fusing it with standard spatio--temporal encoders.

\section{Method}
\label{sec:method}

Our framework consists of two complementary streams. 
(1) A \textbf{backbone stream} utilizes a state-of-the-art skeleton model (\eg GaitMixer~\cite{pinyoanuntapong2023gaitmixer}) to encode global spatio-temporal patterns from joint sequences. 
(2) The proposed \textbf{wavelet feature stream} explicitly models motion dynamics by transforming per-joint velocities into time--frequency representations and learning discriminative dynamic cues (Figure~\ref{fig:wavelet}). 
The two feature vectors are fused and fed to a classifier, yielding higher accuracy and robustness than using either source alone.

\subsection{Wavelet Feature Stream}
\label{sec:wave}
The Continuous Wavelet Transform (CWT) is well suited to \emph{non-stationary} motion signals: by analyzing localized oscillations across scales, it separates stable, periodic gait rhythms from abrupt artifacts (\eg pose jitter or bag swing). A lightweight CNN consumes the CWT maps and produces a compact dynamic descriptor.

\paragraph*{Time--Frequency Representation via CWT.}
\label{sec:cwt}
Given a 2D skeleton sequence \(S\!\in\!\mathbb{R}^{2\times T\times V}\) (axes \(\{x,y\}\), time \(T\), joints \(V\)), we emphasize dynamics by computing per-joint velocities:
\begin{equation}
    v_x(t)=S_x(t)-S_x(t-1),\quad 
    v_y(t)=S_y(t)-S_y(t-1).
\label{eq:vel}
\end{equation}
For each 1D velocity signal \(v(\cdot)\) we apply the CWT with mother wavelet \(\psi\):
\[
\mathcal{W}_\psi v(s,\tau)=\int v(t)\,\psi^*\!\Big(\frac{t-\tau}{s}\Big)\,dt,
\]
where \(s\) and \(\tau\) denote scale and translation. Stacking \(F\) scales yields a time--frequency map (scalogram). After log-magnitude and per-scale normalization, 
the tensor is
\[
H \in \mathbb{R}^{V\times 2\times F\times (T-1)},
\]
which encodes multi-scale dynamics for all joints. 

\paragraph*{Multi-scale CNN for Dynamic Feature Learning.}
\label{sec:cnn}
To learn meaningful features from \(H\), we process each joint/axis map with a small multi-branch 2D CNN over the \((F\times (T{-}1))\) plane:
\begin{itemize}[leftmargin=*]
    \item \emph{Stem:} two sequential convolutions with \(5{\times}5\) then \(3{\times}3\) kernels (each followed by BN--ReLU) to refine local patterns.
    \item \emph{Parallel multi-scale block:} a \(3{\times}3\) branch (fine-grained, subtle changes) and a \(7{\times}7\) branch (coarse rhythm/periodicity) operate concurrently; outputs are concatenated along channels and normalized.
    \item \emph{Pooling:} global average pooling over \((F,t)\) produces a fixed-size vector \(f_{\text{joint}}\in\mathbb{R}^{C_{\text{out}}}\) per joint.
\end{itemize}
Denoting axis-wise features as \(f_{v,x}, f_{v,y}\), we form \(f_v=[f_{v,x};f_{v,y}]\in\mathbb{R}^{2C_{\text{out}}}\). Concatenating all joints and projecting with a fully-connected (FC) layer yields the wavelet descriptor
\[
F_{\text{wavelet}}=\mathrm{FC}\!\left([f_{1};\dots;f_{V}]\right)\in\mathbb{R}^{128}.
\]

\subsection{Feature Fusion and Classification}
\label{sec:fusion}
For the final recognition embedding, we fuse the backbone feature $F_{\text{backbone}}$ and the wavelet descriptor $F_{\text{wavelet}}$ by channel-wise concatenation:
\begin{equation}
    F_{\text{fusion}}=\mathrm{Concat}\!\left(F_{\text{backbone}},\,F_{\text{wavelet}}\right).
\label{eq:fusion}
\end{equation}
The fused vector is then projected to a fixed-dimensional embedding by a fully connected layer:
\begin{equation}    F_{\text{final}}=\mathrm{FC}\!\left(F_{\text{fusion}}\right)\in\mathbb{R}^{128}.
\label{eq:final}
\end{equation}
We apply $\ell_2$ normalization to obtain a unit-norm representation:
\begin{equation}
    \hat F \;=\; \frac{F_{\text{final}}}{\|F_{\text{final}}\|_2}.
\label{eq:norm}
\end{equation}
By injecting explicit time--frequency dynamics, the wavelet stream compensates for subtle motion cues that backbone models may miss, improving robustness under covariate shifts such as carrying bags (BG) or wearing coats (CL).

\subsection{Loss Function}
We use a triplet loss~\cite{schroff2015facenet} to minimize intra-class variation while enlarging inter-class separation. 
Triplets $(a, p, n)$ are constructed with hard mining within each mini-batch: for each anchor $a$, the hardest positive $p$ (farthest sample of the same identity) and the hardest negative $n$ (closest sample of a different identity) are selected to form the triplet set $T$.
With margin $\alpha{=}0.01$, the objective is
\begin{equation}
\begin{split}
L_{\text{tri}}
&= \frac{1}{|T|}\!\sum_{(a,p,n)\in T}
   \max\!\Bigl(0,\; \|\hat F(a)-\hat F(p)\|_2  \\
&\hphantom{=} \qquad\qquad\qquad\qquad\, - \|\hat F(a)-\hat F(n)\|_2 + \alpha\Bigr).
\end{split}
\label{eq:triplet}
\end{equation}
Here, $\hat F(\cdot)$ denotes the $\ell_2$-normalized embedding in \eqref{eq:norm}, making the Euclidean distance equivalent (up to a constant) to cosine distance.

\section{Experiments}
\label{exp}

\subsection{Experimental Settings}

\noindent\textbf{Datasets.}
CASIA-B~\cite{casiab} is a widely used multi-view gait dataset comprising 124 subjects recorded from 11 viewpoints (angles $0^\circ$ to $180^\circ$ in $18^\circ$ steps). 
Each subject provides 10 sequences: six normal walking (NM), two with a coat (CL), and two carrying a bag (BG), totaling 13{,}640 sequences across all views.
Following the standard protocol adopted by prior work~\cite{song2019gaitnet,chao2021gaitset,teepe2021gaitgraph,teepe2022towards,pinyoanuntapong2023gaitmixer}, we use the first 74 subjects for training and the remaining 50 for testing.
For each test subject, NM\#1--4 are used as the gallery, and the remaining six sequences serve as probes: NM\#5--6, BG\#1--2, and CL\#1--2.

\noindent\textbf{Training Details.}
We use HRNet~\cite{Sun2019DeepHR} for 2D pose estimation (same setup as GaitGraph~\cite{teepe2021gaitgraph}), and train with Triplet loss (margin $0.01$) on 60-frame clips. 
Adam with One-Cycle LR~\cite{smith2019super} (init $6\!\times\!10^{-3}$) and weight decay $1\!\times\!10^{-5}$ is used.
A balanced batch sampler ensures an equal number of sequences per identity.
Unless otherwise noted, training runs for 100 epochs and the checkpoint with the best mean Rank-1 accuracy is reported.

\renewcommand{\arraystretch}{1.05}
\begin{table}[b]
\caption{
Comparison with state-of-the-art methods on CASIA-B (Rank-1, \%).
Results are reported under NM, BG, and CL conditions; \emph{Mean} averages the three. 
}
\small
\setlength{\tabcolsep}{3pt}
\centering
\resizebox{\columnwidth}{!}{%
\begin{tabular}{c|c|ccc|c}
    \toprule
    \textbf{Type} & \textbf{Method} & \textbf{NM} & \textbf{BG} & \textbf{CL} & \textbf{Mean} \\ \midrule
    \multirow{4}{*}{\begin{minipage}{2.0cm}\centering Appearance-\\based\end{minipage}} 
     & GaitNet~\cite{song2019gaitnet}    & 91.6 & 85.7 & 58.9 & 78.7 \\
     & GaitSet~\cite{chao2021gaitset}    & 95.0 & 87.2 & 70.4 & 84.2 \\
     & GaitPart~\cite{fan2020gaitpart}   & 96.2 & 91.5 & 78.7 & 88.8 \\ 
     & 3DLocal~\cite{huang20213d}        & \textbf{98.3} & \textbf{95.5} & \textbf{84.5} & \textbf{92.8} \\ \midrule
    \multirow{6}{*}{\begin{minipage}{2.0cm}\centering Skeleton-\\based\end{minipage}} 
     & PoseGait~\cite{liao2020model} & 68.7 & 44.5 & 36.0 & 49.7 \\
     & GaitGraph~\cite{teepe2021gaitgraph} & 87.7 & 74.8 & 66.3 & 76.3 \\
     & GaitGraph2~\cite{teepe2022towards} & 82.0 & 73.2 & 63.6 & 72.9 \\
     & GaitFormer~\cite{pinyoanuntapong2023gaitmixer} & 91.5 & 81.4 & 77.2 & 83.4 \\
     & GaitMixer~\cite{pinyoanuntapong2023gaitmixer} & 94.9 & 85.6 & 84.5 & 88.3 \\ 
     & GaitMixer~\cite{pinyoanuntapong2023gaitmixer} \textbf{+ Ours} & \textbf{95.1} & \textbf{87.2} & \textbf{86.7} & \textbf{89.7} \\ \bottomrule
\end{tabular}
}
\label{tab:appearance_vs_skeleton}
\end{table}

\begin{table*}[t]
\centering
\small
\setlength{\tabcolsep}{7pt}
\caption{
Averaged Rank-1 accuracies (\%) with the wavelet stream on CASIA-B per probe angle (identical-view excluded).
\textbf{Bold} indicates the higher score within each backbone pair (baseline \emph{vs.}\ + Ours).
\underline{Underline} indicates the best score within each condition block (NM/BG/CL).
}
\resizebox{\textwidth}{!}{ 
\begin{tabular}{c|l|ccccccccccc|c}
\toprule
Gallery NM\#1--4 & Method & 0$^\circ$ & 18$^\circ$ & 36$^\circ$ & 54$^\circ$ & 72$^\circ$ & 90$^\circ$ & 108$^\circ$ & 126$^\circ$ & 144$^\circ$ & 162$^\circ$ & 180$^\circ$ & Mean \\
\midrule
\multirow{6}{*}{NM\#5--6} 
& \cellcolor{gray!15}GaitGraph~\cite{teepe2021gaitgraph} 
& \cellcolor{gray!15}85.3 & \cellcolor{gray!15}88.5 & \cellcolor{gray!15}91.0 & \cellcolor{gray!15}\textbf{92.5} & \cellcolor{gray!15}87.2 & \cellcolor{gray!15}86.5 & \cellcolor{gray!15}88.4 & \cellcolor{gray!15}89.2 & \cellcolor{gray!15}87.9 & \cellcolor{gray!15}85.9 & \cellcolor{gray!15}81.9 & \cellcolor{gray!15}87.7 \\
& + Ours 
& \textbf{88.0} & \textbf{91.4} & \textbf{93.0} & 92.3 & \textbf{92.0} & \textbf{88.7} & \textbf{89.0} & \textbf{89.9} & \textbf{92.0} & \textbf{90.2} & \textbf{85.3} & \textbf{90.2} \\ 
& \cellcolor{gray!15}GaitFormer~\cite{pinyoanuntapong2023gaitmixer} 
& \cellcolor{gray!15}90.9 & \cellcolor{gray!15}91.2 & \cellcolor{gray!15}93.7 & \cellcolor{gray!15}91.9 & \cellcolor{gray!15}91.9 & \cellcolor{gray!15}92.7 & \cellcolor{gray!15}93.3 & \cellcolor{gray!15}91.8 & \cellcolor{gray!15}92.5 & \cellcolor{gray!15}90.5 & \cellcolor{gray!15}85.5 & \cellcolor{gray!15}91.5 \\
& + Ours 
& \textbf{92.0} & \textbf{91.8} & \textbf{94.2} & \textbf{94.2} & \textbf{93.0} & \textbf{94.5} & \textbf{94.3} & \textbf{93.4} & \textbf{93.6} & \textbf{91.8} & \textbf{86.7} & \textbf{92.7}\\
& \cellcolor{gray!15}GaitMixer~\cite{pinyoanuntapong2023gaitmixer} 
& \cellcolor{gray!15}\underline{\textbf{94.4}} & \cellcolor{gray!15}\underline{\textbf{94.9}} & \cellcolor{gray!15}94.6 & \cellcolor{gray!15}\underline{\textbf{96.3}} & \cellcolor{gray!15}95.3 & \cellcolor{gray!15}96.3 & \cellcolor{gray!15}95.3 & \cellcolor{gray!15}94.7 & \cellcolor{gray!15}\underline{\textbf{95.3}} & \cellcolor{gray!15}\underline{\textbf{94.7}} & \cellcolor{gray!15}92.2 & \cellcolor{gray!15}94.9 \\
& + Ours 
& \underline{\textbf{94.4}} & 94.4 & \underline{\textbf{95.4}} & \underline{\textbf{96.3}} & \underline{\textbf{95.6}} & \underline{\textbf{96.6}} & \underline{\textbf{95.8}} & \underline{\textbf{95.4}} & 95.0 & 93.6 & \underline{\textbf{93.7}} & \underline{\textbf{95.1}}\\

\midrule
\multirow{6}{*}{BG\#1--2} 
& \cellcolor{gray!15}GaitGraph~\cite{teepe2021gaitgraph} 
& \cellcolor{gray!15}\textbf{75.8} & \cellcolor{gray!15}\textbf{76.7} & \cellcolor{gray!15}75.9 & \cellcolor{gray!15}76.1 & \cellcolor{gray!15}71.4 & \cellcolor{gray!15}73.9 & \cellcolor{gray!15}\textbf{78.0} & \cellcolor{gray!15}\textbf{74.7} & \cellcolor{gray!15}\textbf{75.4} & \cellcolor{gray!15}75.4 & \cellcolor{gray!15}69.2 & \cellcolor{gray!15}74.8 \\
& + Ours 
& 75.2 & 74.6 & \textbf{79.3} & \textbf{80.1} & \textbf{78.9} & \textbf{78.9} & 77.0 & \textbf{78.5} & 75.3 & \textbf{76.9} & \textbf{69.5} & \textbf{76.7} \\ 
& \cellcolor{gray!15}GaitFormer~\cite{pinyoanuntapong2023gaitmixer} 
& \cellcolor{gray!15}\textbf{82.5} & \cellcolor{gray!15}83.2 & \cellcolor{gray!15}85.7 & \cellcolor{gray!15}85.7 & \cellcolor{gray!15}\textbf{84.2} & \cellcolor{gray!15}80.2 & \cellcolor{gray!15}78.9 & \cellcolor{gray!15}82.6 & \cellcolor{gray!15}82.2 & \cellcolor{gray!15}78.6 & \cellcolor{gray!15}71.3 & \cellcolor{gray!15}81.4 \\
& + Ours 
& 81.8 & \textbf{83.6} & \textbf{86.3} & \textbf{86.5} & 83.2 & \textbf{81.9} & \textbf{79.8} & \textbf{83.6} & \textbf{82.4} & \textbf{81.5} & \textbf{73.7} & \textbf{82.2} \\
& \cellcolor{gray!15}GaitMixer~\cite{pinyoanuntapong2023gaitmixer} 
& \cellcolor{gray!15}83.5 & \cellcolor{gray!15}85.6 & \cellcolor{gray!15}\underline{\textbf{88.1}} & \cellcolor{gray!15}89.7 & \cellcolor{gray!15}85.2 & \cellcolor{gray!15}87.4 & \cellcolor{gray!15}84.0 & \cellcolor{gray!15}84.7 & \cellcolor{gray!15}84.6 & \cellcolor{gray!15}87.0 & \cellcolor{gray!15}81.4 & \cellcolor{gray!15}85.6 \\
& + Ours 
& \underline{\textbf{86.7}} & \underline{\textbf{88.2}} & 87.4 & \underline{\textbf{90.0}} & \underline{\textbf{86.4}} & \underline{\textbf{89.0}} & \underline{\textbf{86.2}} & \underline{\textbf{86.1}} & \underline{\textbf{86.9}} & \underline{\textbf{88.3}} & \underline{\textbf{83.6}} & \underline{\textbf{87.2}}\\

\midrule
\multirow{6}{*}{CL\#1--2} 
& \cellcolor{gray!15}GaitGraph~\cite{teepe2021gaitgraph} 
& \cellcolor{gray!15}69.6 & \cellcolor{gray!15}66.1 & \cellcolor{gray!15}68.8 & \cellcolor{gray!15}67.2 & \cellcolor{gray!15}64.5 & \cellcolor{gray!15}62.0 & \cellcolor{gray!15}69.5 & \cellcolor{gray!15}65.6 & \cellcolor{gray!15}65.7 & \cellcolor{gray!15}66.1 & \cellcolor{gray!15}64.3 & \cellcolor{gray!15}66.3 \\
& + Ours 
& \textbf{71.2} & \textbf{70.2} & \textbf{70.4} & \textbf{70.7} & \textbf{66.3} & \textbf{66.5} & \textbf{74.1} & \textbf{69.2} & \textbf{72.0} & \textbf{72.0} & \textbf{67.9} & \textbf{70.0} \\ 
& \cellcolor{gray!15}GaitFormer~\cite{pinyoanuntapong2023gaitmixer} 
& \cellcolor{gray!15}76.1 & \cellcolor{gray!15}80.3 & \cellcolor{gray!15}81.0 & \cellcolor{gray!15}78.2 & \cellcolor{gray!15}77.7 & \cellcolor{gray!15}76.6 & \cellcolor{gray!15}77.4 & \cellcolor{gray!15}75.8 & \cellcolor{gray!15}76.5 & \cellcolor{gray!15}75.7 & \cellcolor{gray!15}\textbf{77.2} & \cellcolor{gray!15}77.2 \\
& + Ours 
&  \textbf{78.6} & \textbf{81.6} & \textbf{82.8} & \textbf{81.3} & \textbf{79.9} & \textbf{77.3} & \textbf{80.3} & \textbf{78.8} & \textbf{78.2} & \textbf{79.2} & 75.1 & \textbf{79.4}\\
& \cellcolor{gray!15}GaitMixer~\cite{pinyoanuntapong2023gaitmixer} 
& \cellcolor{gray!15}81.2 & \cellcolor{gray!15}83.6 & \cellcolor{gray!15}82.3 & \cellcolor{gray!15}83.5 & \cellcolor{gray!15}84.5 & \cellcolor{gray!15}84.8 & \cellcolor{gray!15}86.9 & \cellcolor{gray!15}88.9 & \cellcolor{gray!15}87.0 & \cellcolor{gray!15}85.7 & \cellcolor{gray!15}81.6 & \cellcolor{gray!15}84.5 \\
& + Ours 
& \underline{\textbf{84.0}} & \underline{\textbf{86.3}} & \underline{\textbf{83.9}} & \underline{\textbf{87.5}} & \underline{\textbf{86.8}} & \underline{\textbf{86.9}} & \underline{\textbf{89.3}} & \underline{\textbf{89.1}} & \underline{\textbf{87.4}} & \underline{\textbf{88.5}} & \underline{\textbf{83.9}} & \underline{\textbf{86.7}} \\
\bottomrule
\end{tabular}}
\label{tab:skeletonSOTA}
\end{table*}

\begin{table}[t]
\centering
\caption{Ablation on the mother wavelet for CWT on CASIA-B (Rank-1, \%). 
}
\label{tab:wavelet_ablation}
\small
\setlength{\tabcolsep}{10pt}
\centering
\resizebox{\columnwidth}{!}{%
\begin{tabular}{l|ccc|c}
\toprule
\textbf{Wavelet} & \textbf{NM} & \textbf{BG} & \textbf{CL} & \textbf{Mean} \\
\midrule
Mexican Hat (\texttt{mexh})& 94.7 & 86.3 & 85.7 & 88.8 \\
Gaussian (\texttt{gaus1})  & 95.1 & 86.4 & 84.9 & 88.8 \\
Shannon (\texttt{shan})    & 94.9 & 86.1 & 84.4 & 88.5 \\
Morlet (\texttt{morl})     & \textbf{95.1} & \textbf{87.2} & \textbf{86.7} & \textbf{89.7} \\
\bottomrule
\end{tabular}
}
\end{table}

\subsection{Comparison with Prior Art}
We compare against representative \emph{appearance-based} and \emph{skeleton-based} methods on CASIA-B (Table~\ref{tab:appearance_vs_skeleton}; Rank-1). 
Our approach is skeleton-based: attaching the wavelet stream to GaitMixer yields a new SOTA among skeleton methods and maintains strong performance under covariates. 
Notably, under the coat condition (CL) -- the most challenging setting -- our model achieves the best accuracy across \emph{all} methods (including appearance-based), highlighting robustness to appearance changes.
As expected, the appearance-based 3DLocal remains strongest in mean over all conditions; nevertheless, our module consistently lifts the best skeleton baseline and narrows the overall gap, underscoring the complementarity of explicit time--frequency modeling.

\subsection{Plug-and-Play Gains Across Backbones}
We attach the proposed wavelet stream to multiple skeleton backbones (GaitGraph~\cite{teepe2021gaitgraph}, GaitFormer, GaitMixer~\cite{pinyoanuntapong2023gaitmixer}) and evaluate \emph{per probe angle} excluding identical-view cases (Table~\ref{tab:skeletonSOTA}). 
The largest and most consistent gains appear under the coat condition (CL): mean CL accuracy improves by +3.7 pp on GaitGraph (66.3$\to$70.0), +2.2 pp on GaitFormer (77.2$\to$79.4), and +2.2 pp on GaitMixer (84.5$\to$86.7).
For NM and BG, we observe the same trend -- smaller but steady improvements across angles and backbones -- confirming that explicit time--frequency dynamics provide complementary cues to standard spatio--temporal encoders.

\subsection{Analysis on the Choice of Wavelet Function}
Table~\ref{tab:wavelet_ablation} summarizes the ablation on different mother wavelets. 
Morlet yields the highest mean accuracy (89.7\%) and strong results across NM/BG/CL. 
We attribute this to its favorable joint time--frequency localization and complex (analytic) form, which capture both amplitude and phase of quasi-periodic gait oscillations. 
In contrast, Mexican Hat and Gaussian wavelets are real and derivative-like, emphasizing localized transients/edges; they are slightly less effective for modeling cycle-level rhythmic dynamics. 
Shannon exhibits ideal frequency localization but poor temporal compactness (sinc-like behavior), which can blur short-term structure and leads to the lowest mean among the tested options. 
Unless otherwise noted, we therefore adopt \emph{Morlet} as the default wavelet in all experiments.

\section{Conclusion}
\label{sec:conclusion}
We presented a plug-and-play \emph{Wavelet Feature Stream} that injects explicit time--frequency dynamics of joint velocities into skeleton-based gait recognition.
By transforming per-joint velocities with the continuous wavelet transform and learning multi-scale patterns with a lightweight CNN, our module complements spatial modeling in standard backbones without architectural changes.
On CASIA-B, attaching the stream to GaitMixer establishes a new skeleton-based SOTA and yields consistent gains across multiple backbones; the improvements are particularly pronounced under covariate shifts such as carrying bags (BG) and wearing coats (CL), where our model also surpasses leading appearance-based methods in the CL setting.
Future work includes learning wavelet parameters end-to-end, extending to cross-dataset and cross-sensor scenarios (\eg IMU), and exploring adaptive fusion with backbone features.


\noindent\textbf{Acknowledgements.}
This work was supported by the National Research Foundation of Korea (NRF) grant funded by the Korea government (MSIT) (No. RS-2025-16070597) and Global - Learning \& Academic research institution for Master’s·PhD students, and Postdocs (G-LAMP) Program of the National Research Foundation of Korea (NRF) grant funded by the Ministry of Education (No. RS-2025-25442252).

\bibliographystyle{IEEEbib}
\bibliography{refs}

\end{document}